%% file: main.tex
\documentclass[a4paper,twoside]{article}

\usepackage[utf8]{inputenc}
\usepackage[T1]{fontenc}
\usepackage[american]{babel}

\usepackage{tikz}
\usetikzlibrary{shapes,arrows,automata,plotmarks,positioning}

\usepackage{epsfig}
\usepackage{subcaption}
\usepackage{calc}
\usepackage{amssymb}
\usepackage{amstext}
\usepackage{amsmath}
\usepackage{amsthm}
\usepackage{multicol}
\usepackage{pslatex}
\usepackage{apalike}
\usepackage{siunitx}
\usepackage{booktabs}
\usepackage{multirow}
\usepackage{hyperref}
\usepackage[inline]{enumitem}
\usepackage[capitalise,noabbrev]{cleveref}
\usepackage{balance}

\usepackage{SCITEPRESS}

\input{macros}

\begin{document}

\title{Anomaly Detection in Beehives using Deep Recurrent Autoencoders}

\author{
\authorname{
Padraig Davidson\sup{1},
Michael Steininger\sup{1},
Florian Lautenschlager\sup{1},
Konstantin Kobs\sup{1},
Anna Krause\sup{1}
and Andreas Hotho\sup{1}}
\affiliation{\sup{1}Institute of Computer Science, Chair of Computer Science X, University of Würzburg, Am Hubland, Würzburg, Germany}
\email{\{davidson, steininger, lautenschlager, kobs, anna.krause, hotho\}@informatik.uni-wuerzburg.de}
}


\abstract{Precision beekeeping allows to monitor bees' living conditions by equipping beehives with sensors.
The data recorded by these hives can be analyzed by machine learning models to learn behavioral patterns of or search for unusual events in bee colonies.
One typical target is the early detection of bee swarming as apiarists want to avoid this due to economical reasons.
Advanced methods should be able to detect any other unusual or abnormal behavior arising from illness of bees or from technical reasons, e.g. sensor failure.
\newline 
In this position paper we present an autoencoder, a deep learning model, which detects any type of anomaly in data independent of its origin.
Our model is able to reveal the same swarms as a simple rule-based swarm detection algorithm but is also triggered by any other anomaly.
We evaluated our model on real world data sets that were collected on different hives and with different sensor setups.
} % Currently: ~140/200 words

\onecolumn \maketitle \normalsize \setcounter{footnote}{0} \vfill

%\todo{Page limit: 4 < \#pages < 9, preferred 6-7, including references}
\section{\uppercase{Introduction}}
\label{sec:introduction}
\input{sections/introduction.tex}

\section{\uppercase{Related Work}}
\label{sec:rel_work}
\input{sections/related_work.tex}

\section{\uppercase{Datasets}}
\label{sec:dataset}
\input{sections/dataset.tex}

\section{\uppercase{Autoencoder}}
\label{sec:methods}
\input{sections/methods.tex}

\section{\uppercase{Experimental Setup}}
\label{sec:experiments}
\input{sections/experiments.tex}

\section{\uppercase{Results}}
\label{sec:results}
\input{sections/results.tex}

\section{\uppercase{Discussion}}
\label{sec:discussion}
\input{sections/analysis.tex}

\input{sections/discussion.tex}

\section{\uppercase{Conclusion/Future Work}}
\label{sec:conclusion}
\input{sections/conclusion.tex}

\section*{\uppercase{Acknowledgements}}
\input{sections/acknowledgements.tex}

\balance
\bibliographystyle{apalike}
{\small\bibliography{bibliography}}

%\clearpage
%\onecolumn
%\input{sections/appendix.tex}

\end{document}

%% file: macros.tex
\newcommand{\wue}{Würzburg}
\newcommand{\shw}{Bad Schwartau}
\newcommand{\ind}{Markt Indersdorf}
\newcommand{\hobos}{HOBOS}
\newcommand{\webee}{we4bee}
\newcommand{\jel}{Jelgava}
\newcommand{\algo}{RBA}

\newcommand{\para}[1]{\smallskip\noindent\textbf{#1}.}

%% file: sections/introduction.tex
\noindent
%1) Precision beekeeping + anomaly types
Precision apiculture, also known as precision beekeeping, aims to support beekeepers in their care decisions to maximize efficiency.
For that, sensor data is collected on
\begin{itemize*}
    \item[1)] apiary-level (e.g. meteorological parameters),
    \item[2)] colony-level (e.g. beehive temperature), or
    \item[3)] individual bee-related level (e.g. bee counter)
\end{itemize*}
~\cite{zacepins2015challenges}.
To gather data on colony level, beehives are equipped with environmental sensors that continuously monitor and quantify the beehive's state.
%These datasets contain only few events that require immediate attention.
Occasionally there are sensor readings that deviate substantially from the norm.
We refer to these events as anomalies.
They can be categorized as behavior anomalies, sensor anomalies, and external interferences.
The first type describes irregular behavior within the monitored subject, the second describes any abnormal measurements of the used sensors, whereas the last can be subsumed as any exterior force operating.

%2) Explanation of behavioral anomalies
A prominent behavioral anomaly in apiculture is swarming.
Swarming is the event of a colony's queen leaving the hive with a party of worker bees to start a new colony in a distant location.
It is a naturally occurring, albeit highly stochastic reproduction process in a beehive.
%During the \textit{prime swarm} around two thirds of the colony leave the current beehive.
During the \textit{prime swarm} the current queen departs with many of the workers from the old colony.
Subsequent \textit{after swarms} can occur with fewer workers leaving the hive.
Swarming events can reoccur until the total depletion of the original colony~\cite{winston1980swarming}.
%colony~\cite{avitabile1975pheromones,winston1980swarming,villa2004swarming-behavior}
Swarming diminishes a beehive's production and requires the beekeeper's immediate attention, if new colonies are to be recollected.
Therefore, beekeepers try to prevent swarming events in their beehives. % are disadvantageous and are tried to be prevented.

%Swarming is not the only behavioral anomaly that can arise in beehives.
A second notable behavioral anomaly in beehives are mite infestations (\textit{varroa destructor})~\cite{navajas2008differential}.
%~\cite{kralj2007parasitic,navajas2008differential}.
They weaken colonies and make bees more susceptible to additional diseases.
Over time, they lead to bee deaths and thus have severe consequences to the environment as they reduce the pollination power of bees in general.
Just like swarming, mite infestations and diseases require immediate attention by beekeepers.
%Therefore it is important to monitor the beehive's state to timely detect these deviations and react accordingly, e.g. using powdered sugar to tackle mite infestation.

%3) Explanation of sensor anomalies
Sensor anomalies and external interferences are non-bee-related anomalies. They can arise in any sensor network.
Any technical defects including faulty sensors can be summarized as sensor anomalies and require maintenance of the beehive and sensor network.
%4) Explanation of intrusions
In apiaries, external interference is physical interaction of beekeepers, or other external forces, with their hives, e.g. opening the beehive for honey yield.
%could be the physical change of the beehive due to honey yield or other inspection of the colony.
%Precision beekeeping aims to support beekeepers in their actions.
%It requires beehives that are continuously monitored with environmental sensors to capture and quantify the beehive's state.
%These sensor readings bear the potential to detect behavioral anomalies early.
%However, these electronic devices incur a second type of anomaly: sensor anomalies.

%5) DL
%Finding anomalies within these sensor networks is crucial, since they can have a huge economical impact.
Finding anomalies in large datasets requires specialized methods that extend beyond manual evaluations.
Autoencoders (AEs) are a popular choice in anomaly detection.
They are a deep neural network architecture, that is designed to reconstruct normal behavior with minimal loss of information.
In contrast, an AE's reconstruction of anomalous behavior shows significant loss and can therefore be identified.
They are purely data-driven without the need for beehive specific knowledge.

%6) Our Work
In this paper, we present an autoencoder that can detect all three types of anomalies in beehive data in a data-driven fashion.
Our contribution is twofold:
First, we explore the possibility of using an autoencoder for anomaly detection on beehive data.
Second, we show that this architecture can be applied to different types of beehives due to its data-driven origin, without the need for additional fine-tuning.

We evaluate our approach on three datasets:
One long term dataset of four years provided by the \hobos\ (\url{https://hobos.de/}) project, one short term dataset obtained from~\cite{zacepins2016remote}, and another short term dataset from \webee\ (\url{https://we4bee.org/}).
For this preliminary study, we focus on time spans where swarming can occur to show that our approach is working in general.

The remainder of this paper is structured as follows:
% After this introduction we present an outline on past research done in the field of precision beekeeping with special focus on the detection of swarming events.
\cref{sec:rel_work} presents related research. % done in the field of swarm detection. % and anomaly detection techniques for sequential data.
%\cref{sec:dataset} describes the used datasets in detail, whereas \cref{sec:methods} concentrates on a comprehensive characterization of the methods, especially the used autoencoder network structure.
\cref{sec:dataset} describes the used datasets in detail, whereas \cref{sec:methods} concentrates on a comprehensive characterization of our autoencoder network structure.
In \cref{sec:experiments} we describe our experiments and list the results on swarming data in \cref{sec:results}.
\cref{sec:discussion} investigates normal behavior in contrast to selected anomalies, the sensor foundation to detect those with our model, and discusses the results.
We conclude with a summary and possible directions for future work in \cref{sec:conclusion}.

%Note:
%- Motivation: Bienensterben, etc
%- Selling Point: Anomalie Detection (Swarming, strange T stuff) --> Sensoranomalie und Verhaltensanomalie
%- Methode detailieren
%- Beschreibung: Schwärmen geht; an einigen Stellen geht es noch nicht -> Future work: Klären mit Biologen;
%- Future Work: evntl. weitere Experimente
%- Thresholdbestimmung? An Trainingsdaten, Validationsdaten? Zweiter (händischer) Schritt? --> Label?
%- Methode ist invariant zum Bienenstock und Datengetrieben

%Note 2: Geordnet nach Wichtigkeit!
%0) You can do it! :)
%1) Schreiben
%2) Glatt ziehen
%3) Nochmal was Labeln und Experimente machen (wenn Schreiben nicht mehr geht)

%% file: sections/related_work.tex
\noindent
In order to learn how to distinguish normal from anomalous behavior, we use anomaly detection techniques based on neural networks.
In particular, we use recurrent autoencoders, since they have shown to work in many anomaly detection settings with sequential data before~\cite{filonov2016multivariate,malhotra2016multi,shipmon2017time,chalapathy2019deepad}.
Prior work for anomaly detection in bee data has focused mainly on swarm detection, for which several techniques have been published.

\cite{ferrari2008monitoring} monitored sound, temperature and humidity of three beehives to investigate changes of these variables during swarming.
The beehives experienced nine swarming activities during the monitoring period, for which they analyzed the collected data.
They concluded that the shift in sound frequency and the change in temperature might be used to predict swarming.

% \cite{kridi2014predictive} proposed a clustering-based approach to identify pre-swarming behavior.
% Typical daily temperature patterns were obtained through clustering.
\cite{kridi2014predictive} proposed an approach to identify pre-swarming behavior by clustering temperatures into typical daily patterns.
An anomaly is detected if the measurements do not fit into the typical clusters for multiple hours.

\cite{zacepins2016remote} used a customized swarming detection algorithm based on single-point temperature monitoring.
They asserted a base temperature of \SI{34.5}{\celsius} within the hive, which is allowed to fluctuate within $\pm$ \SI{1}{\celsius}.
If an increase of $\geq$ \SI{1}{\celsius} lasted between two and twenty minutes, they reported the timestamp of the peak temperature as the swarming time.

\cite{zhu2019increase} found, that a linear temperature increase can be observed before swarming.
They proposed to measure the temperature between the wall of the hive and the first frame near the bottom which provides the most apparent temperature increase.

While swarming is an important type of anomaly, we believe that other exceptional events should also be detected.

%% file: sections/dataset.tex
\noindent
We use sensor measurements from \hobos, a subset of the data from \webee\ and the data used by~\cite{zacepins2016remote} (referred to as \jel) in our work.
All datasets are referenced by the location of the beehives.

\subsection{\wue\ \& \shw}
\hobos\ equipped five beehives (species: \textit{apis mellifera}; beehive type: zander beehive) with several environmental sensors.
We use data from two beehives, located in \shw\ and \wue.
%which is why this beehive was chosen.
While there are three verified swarming events at \shw, the \wue\ beehive data is completely unlabeled.
We use this beehive to assess cross-beehive applicability of our model.
\begin{figure}
    \centering
    \includegraphics[width=0.9\columnwidth]{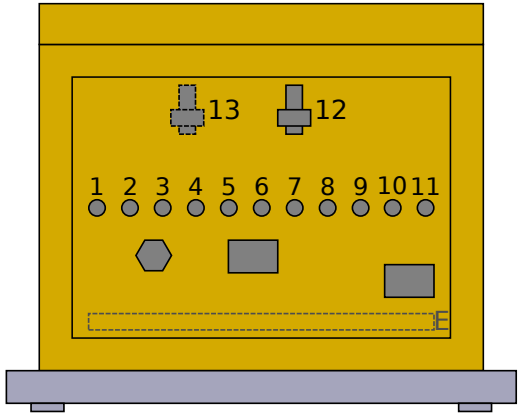}
    \caption{Back of a \hobos\ beehive.
    Temperature sensors T$_1$--T$_{11}$ are mounted between honeycombs, temperature sensors T$_{12}$ and T$_{13}$ are mounted on the back and the front of the hive, respectively.
    E denotes the hive's entrance on the front of the beehive.}
    \label{fig:hobos_sensors}
\end{figure}
\cref{fig:hobos_sensors} shows the back of a \hobos~beehive with all 13 temperature sensors.
The beehive in \shw\ is not equipped with T$_2$, T$_3$, T$_9$, T$_{10}$ and T$_{12}$ while the one in \wue\ has all sensors except T$_2$ and T$_3$.
Additionally, weight, humidity and carbon dioxide (CO$_2$) are measured in the beehives.
Measurements are taken once a minute at every sensor.
Data was collected from May 2016 through September 2019.
As we are mainly interested in swarming events, we only used the data from the typical swarming period May to September of each year for this preliminary study~\cite{fell1977seasonal}.
\hobos\ granted us access to their complete dataset.

\subsection{\jel}
Ten colonies (\textit{apis mellifera mellifera}; norwegian-type hive bodies) were monitored by a single temperature sensor placed above the hive body.
Measurements were recorded once every minute over the time span of May through August in 2015.
The authors granted us access to the nine days in their dataset which contain swarming events, one each day.

\subsection{\ind}
\begin{figure}
    \centering
    \includegraphics[width=0.9\columnwidth]{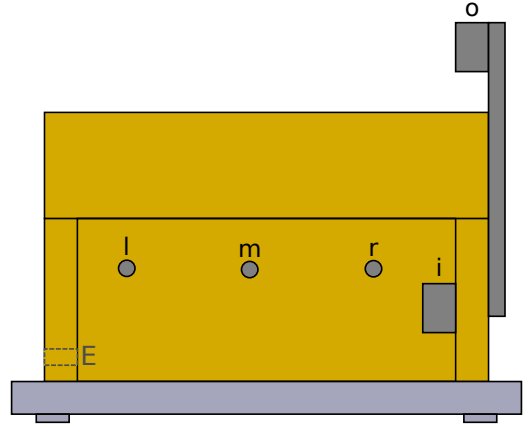}
    \caption{Cutaway view of a \webee\ beehive.
    T$_l$, T$_m$, T$_r$, and T$_i$ are mounted on the inside, laterally to the honeycombs.
    T$_o$ is placed outside at the pylon.
    E denotes the entrance on the front of the beehive.}
    \label{fig:we4bee_sensors}
\end{figure}

The colony (\textit{apis mellifera}; top bar hive) in \ind\ is monitored by five temperature sensors: one on the outside and four on the inside of the beehive.
Three temperature sensors measure laterally to the orientation of the top bars, the remaining one on the inside is placed in parallel at the back.
%T$_l$ is the sensor closest to the entrance hole, T$_m$ and T$_r$ follow subsequently with equidistant space towards the back of the hive.
\cref{fig:we4bee_sensors} shows a cutaway view of a \webee\ hive.
Additionally other environmental influences are monitored with sensors for air pressure, weight, fine dust, humidity, rain and wind.
%Measurements are taken once every second, except for fine dust, which is measured every three minutes.
Measurements are taken once every second (fine dust: every three minutes).
Data ranges from June (start of the colony) through September 2019.

%% file: sections/methods.tex
\noindent
Since autoencoders (AE) have proven to be successful for anomaly detection, we use such a model for our task~\cite{goodfellow2016deeplearning,sakurada2014aedetection,chalapathy2019deepad}.
Especially for analyzing anomalies within sequential data, e.g. time series, deep recurrent autoencoders using long-short-term memory networks (LSTMs)~\cite{goodfellow2016deeplearning} have shown great success over conventional methods (e.g. SVM)~\cite{ergen2017unsupervised}.
We adapt this model for this work.

An autoencoder is a pair of neural networks: an encoder $\phi: \mathcal{X} \rightarrow \mathcal{F}$ and a decoder $\psi: \mathcal{F} \rightarrow \mathcal{X}$, where $\mathcal{X}$ is the input space and $\mathcal{F}$ is the feature space or latent space.
In the case of deep recurrent autoencoders, both, encoder and decoder are LSTMs.
The training objective of the autoencoder is to reconstruct the input: $\bar{x} = \psi(\phi(x)) \sim x$.
Usually, the latent space is smaller than the input space, producing a bottleneck that forces the autoencoder to encode patterns of the input data distribution in the encoder's and decoder's weights.
During the training phase, the autoencoder is provided with data of normal behavior.
%Accordingly, the model learns to reconstruct normal data well.
It learns to reconstruct normal data well.
When it encounters abnormal data during inference, i.e. input data that does not fit into the learned patterns, it is not able to reconstruct the input properly.
Large deviations between model output and data indicate anomalies.

% More formally, the autoencoder tunes the two networks such that they minimize a given reconstruction loss function:
More formally, the two networks are tuned to minimize a given reconstruction loss function:
\[ \phi, \psi = \text{arg min}_{\phi, \psi} \mathcal{L}(x - \psi(\phi(x))). \]
Commonly the $l_2$ norm~\cite{zhou2017anomaly} or the Mean-Squared-Error (MSE)~\cite{shipmon2017time} are chosen as $\mathcal{L}$.
If this loss is greater than a given threshold $\alpha$, the input is considered an anomaly:
\[ \mathcal{L}(x - \psi(\phi(x))) = \mathcal{L}(x - \bar{x}) \geq \alpha \]
$\alpha$ can be set manually or based on a validation dataset containing anomalies, depending on the desired sensibility of the model.
It should ideally be chosen in such a way that all validation anomalies are detected and no normal behavior is misclassified.
%During the training phase, the autoencoder learns to reconstruct normal data as well as possible.
%This implies, that training and validation data should only contain such time series.
%In general, the autoencoder learns to compress normal data very well, while it fails to extract needed indicators for compressing anomalies.
%That entails a high reconstruction error for anomalous data.

%% file: sections/experiments.tex
\noindent
We evaluate our AE model on beehive data from the four locations described in \cref{sec:dataset}.

\begin{figure}
    \centering
    
    \resizebox{\columnwidth}{!}{
        \begin{tikzpicture}
            \usetikzlibrary{calc}
            \usetikzlibrary{decorations.pathreplacing}
            \tikzstyle{block}=[rectangle,draw,align=center,minimum height=2em, outer sep=0cm]
            \tikzstyle{train}=[text width=7.5em]
            \tikzstyle{val}=[text width=4.5em]
            \tikzstyle{test}=[text width=4.5em]
            
            \tikzstyle{nameblock}=[rectangle,align=right,text width=7em,minimum height=2em]
            \tikzstyle{labelblock}=[rectangle,align=center,text width=7em,minimum height=2em]
            \tikzstyle{greycolor}=[draw=black!25, color=black!25]
        
            \node[block, test] (WueAnomaly) {Holdout\\\small{\SI{69132}{\minute}}};
            \node[block, val, left=0.2cm of WueAnomaly] (WueVal) {Validation\\\small{\SI{75314}{\minute}}};
            \node[block, train, left=0cm of WueVal] (WueTrain) {Training\\\small{\SI{677822}{\minute}}};
            \node[nameblock, left=0.2cm of WueTrain] (Wue) {\textbf{\shw}};
            
            \node[block, test, below=0.5cm of WueAnomaly] (BadSAnomaly) {Test\\\small{\SI{72013}{\minute}}};
            \node[block, val, greycolor, left=0.2cm of BadSAnomaly] (BadSVal) {Validation\\\small{\SI{46082}{\minute}}};
            \node[block, train, greycolor, left=0cm of BadSVal] (BadSTrain) {Training\\\small{\SI{42134}{\minute}}};
            \node[nameblock, left=0.2cm of BadSTrain] (BadS) {\wue};
            
            \node[block, test, below=0.5cm of BadSAnomaly] (JelAnomaly) {Test\\\small{\SI{12960}{\minute}}};
            \node[nameblock, below=0.5cm of BadS] (Jel) {\jel};
            
            \node[block, test, below=0.5cm of JelAnomaly] (IndAnomaly) {Test\\\small{\SI{100021}{\minute}}};
            \node[nameblock, below=0.5cm of Jel] (Ind) {\ind};
            
            % Labels above
            % \node[labelblock, above of=WueAnomaly] (LabelAnomaly) {\textbf{Anomalous Behavior}};
            % \node[labelblock, left=0.2cm of LabelAnomaly, above of=WueTrain, fit=(WueTrain)(WueVal)] (LabelNormal) {\textbf{Normal Behavior}};
            \draw [decorate,decoration={brace,amplitude=10pt, raise=0.1cm}] (WueTrain.north west) -- (WueVal.north east) node [midway,above=0.6cm] {Normal Behavior};
            \draw [decorate,decoration={brace,amplitude=10pt, raise=0.1cm}] (WueAnomaly.north west) -- (WueAnomaly.north east) node [midway,above=0.6cm] {Anomalous Behavior};
            
        \end{tikzpicture}
    }
    
    \caption{The data splits used for \shw. The autoencoder is trained on \shw's `Training'.
    The hyperparameters and \(\alpha\) are tuned using its `Validation' and `Holdout', respectively.
    The model is then tested on all `Test'.
    For \wue, the splits are set accordingly using its `Training', `Validation', and `Test' as `Holdout'.
    We provide the recording time for all splits.
    }
    \label{fig:data_splitting}
\end{figure}
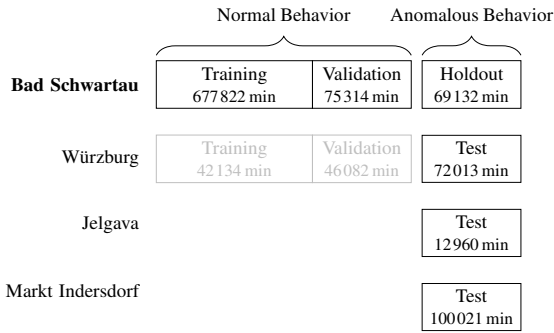

\para{Data splitting}
Both \hobos\ hives, \shw\ and \wue, were used for training purposes.
Through visual analysis of the data, we labeled each day of the dataset as either normal or anomalous.
According to~\cite{zacepins2016remote,ferrari2008monitoring} a fully enlarged colony maintains a constant core temperature of \SI{34.5}{\celsius}.
All days with much higher or lower temperature readings were considered to contain anomalies.
%separated the dataset into normal and abnormal behavior as well as possible.
The training and validation set is sampled from the normal portion of the dataset.
The holdout set contains all days with abnormal behavior from the beehive used for training, while the test set contains all days with anomalies from any other beehive.
\cref{fig:data_splitting} visualizes the procedure exemplary when training on the \shw\ hive.
Days labeled as abnormal typically also contain fragments of normal behavior.
This implies that test and holdout sets are a mixture of normal and anomalous behavior.

%Additionally, we evaluated all models on the \jel\ dataset in order to maintain comparability regarding the potential of swarm detection, and the \ind\ dataset for other anomalies.
%The beehive in \ind\ was also only used for testing purposes, since it is short term and unlabeled.

%\cref{tab:dataset_description} provides a detailed overview of the data split used for training and testing.
% \cref{tab:dataset_description} provides a detailed overview over the dataset, implicitly showing the split used for training and testing.
% \begin{table}[t]
%     \centering
%     \caption{Dataset description. The first column displays beehive's location, the second one the overall number of measurements. The last column lists the number of available temperature sensors in the beehive. (N) stands for normal, (A) for anomalous and (S) for swarming data.}
%     \begin{tabular}{@{}lrr@{}}
%         \toprule
%         Beehive &  {Measurements} & |T|\\
%         \midrule
%         \shw~(N) & \num{753136} & \multirow{2}{*}{8} \\
%         \shw~(A) & \num{69132} & \\
%         \midrule
%         \wue~(N) & \num{460816} & \multirow{2}{*}{12} \\
%         \wue~(A) & \num{72013} & \\
%         \midrule
%         \jel~(S) & \num{12960} & 1\\
%         \midrule
%         \ind~(A) & \num{100021} & 5\\
%         \bottomrule
%     \end{tabular}
%     \label{tab:dataset_description}
% \end{table}

\para{Input data}
We use the centrally located temperature sensor T$_6$ for the locations \wue\ and \shw, in \ind\ we use T$_m$, downsampling its measurements to one minute resolution.
In \jel\ the only temperature sensor available is utilized.
We have also tried T$_r$ for \ind\ and T$_8$ for \wue\ and \shw\ in another experiment to analyze differences in predictions when using other sensors.

The temperature data is provided to the model in windows containing a fixed number of subsequent measurements.
We used a window size of \SI{60}{\minute}, i.e. 60 measurements, since a swarming event usually ranges from \SI{20}{\minute} to \SI{60}{\minute}~\cite{zhu2019increase,zacepins2016remote,ferrari2008monitoring}.
For augmentation purposes we built all possible windows of consecutive measurements.
All time series were normalized by their z-score.

\para{Model training}
We performed a random grid search~\cite{bergstra2012random} to find the best hyperparameters, i.e. the hidden size $hs \in \{2,\ldots,64\}$ and the number of layers $n \in \{1,\ldots,4\}$ for the encoder and decoder LSTMs.

For training we used the \textit{Adam} optimizer~\cite{kingma2014adam} with the default parameters (learning rate $10^{-3}$) and MSE as the loss function.
To prevent overfitting, we employed early stopping with a patience of five epochs and used a ten percent split of the training data for validation.

As described in \cref{sec:methods}, any time series with a reconstruction error larger than $\alpha$ is considered an anomaly.
We selected the threshold manually by examining plots of the found anomalies and gradually decreased this value so that no false positive was detected in the holdout set.
As a holdout set we used the test set from the same beehive as the autoencoder was trained on (see \cref{fig:data_splitting}).

%\subsection{Predictions}
\para{Predictions}
The rule-based algorithm described in~\cite{zacepins2016remote} (referred to as \algo) was used on all subsets.
That is, it was used on the training and testing data from the colonies in \shw, \wue\ and \ind, as well as the one in \jel\ itself.
We found no swarming events, neither false nor true positives, with this method in any training set, verifying our manual selection of training data.
Where possible, we tested it with several temperature sensors.

We trained the AE on both \hobos\ hives independently and used their respective holdout set for setting the anomaly threshold.
After that, we used this threshold and the trained model as an inference model to predict anomalies in all other anomaly sets, e.g. \shw\ was used to predict anomalies in \wue, \jel\ and \ind.
\jel\ and \ind\ were not used for training purposes, since the former only contains anomalous data, and the latter was installed only recently.

%% file: sections/results.tex
\begin{figure}
    \captionsetup[subfigure]{skip=-5pt}
    \centering
    \begin{subfigure}{\columnwidth}
        \centering
        \includegraphics[width=\textwidth]{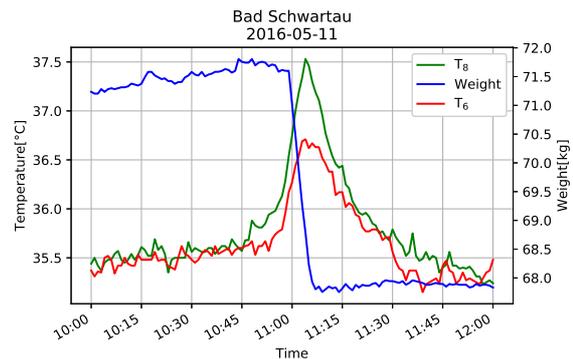}
        \caption{(Prototypical) Swarm as indicated by T$_6$ and T$_8$, detected by \algo\ and AE.}
        \label{fig:colony_behavior_swarm_real_detected}
    \end{subfigure}%

    \begin{subfigure}[t]{\columnwidth}
        \centering
        \includegraphics[width=\textwidth]{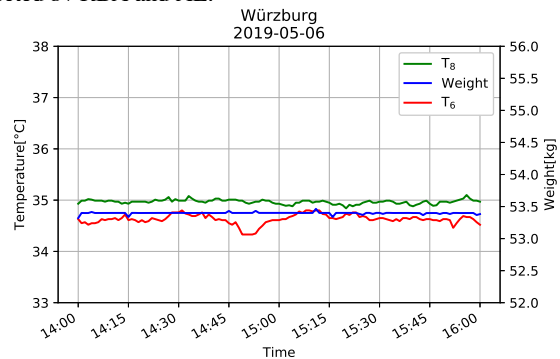}
        \caption{Normal behavior of all three sensors}
        \label{fig:colony_behavior_normal}
    \end{subfigure}%

    \caption{Exemplary data.
    (\ref{fig:colony_behavior_swarm_real_detected}) Expected variations of all sensors for a swarm. (\ref{fig:colony_behavior_normal}) Expected variations of all sensors for normal behavior.
    }
    \label{fig:swam_and_normal}
\end{figure}

\noindent
\cref{tab:results_predictions} lists all known or found swarming events using the temperature sensor T$_6$ or T$_m$.
\begin{table}
    \vspace{5pt}
    \centering
    \caption{Detected Anomalies. The first column shows the name of the used test (anomaly) set.
    (S) signifies that the set contains swarms while (O) stands for other anomalies.
    The next column displays the date of the event, and --- where suitable --- a reference to subfigures in \cref{fig:swam_and_normal,fig:colony_behavior,fig:colony_behavior2}.
    %The last two columns indicate whether the rule-based algorithm proposed by~\cite{zacepins2016remote} (RB) or our method (AE) detected the anomaly.
    The last two columns indicate whether \algo\ or our method (AE) detected the anomaly.
    Predictions on \hobos-hives are based on sensor T$_6$, on T$_m$ for \webee.
    We used the \shw\ trained model to predict the swarms in any other beehive, except for \shw\ itself.}
    \resizebox{\columnwidth}{!}{
    \begin{tabular}{@{}llcc@{}}
        \toprule
        \multirow{2}{*}{Dataset} & \multirow{2}{*}{Timestamp} & \multicolumn{2}{c}{Detected}\\
        \cmidrule{3-4}
        % &&&\cite{zacepins2016remote} & AE\\
        && \algo & AE\\
        \midrule
        \multirow{7}{*}{\shw~(S)} & 2016-05-11 11:05\textsuperscript{\ref{fig:colony_behavior_swarm_real_detected}} & $\checkmark$ & $\checkmark$\\
         & 2016-05-22 07:30 & $\checkmark$ & $\checkmark$\\
         & 2017-06-06 15:02 & $\checkmark$ & $\checkmark$\\
         & 2019-05-13 09:30$^\star$ & $\checkmark$ & $\checkmark$\\
         & 2019-05-21 09:15$^\star$ & $\checkmark$ & $\checkmark$\\
         & 2019-05-25 12:00$^\star$ & $\checkmark$ & $\checkmark$\\
         \cmidrule{2-4}
        \shw~(O) & 2016-08-03 17:24 & $\checkmark$ & $\checkmark$\\
        \midrule
        \multirow{2}{*}{\wue~(S)} & 2019-05-01 09:15\textsuperscript{\ref{fig:colony_behavior_swarm_not_detected}} & & $\checkmark$\\
        & 2019-05-10 11:15\textsuperscript{\ref{fig:colony_behavior_triple_swarm}} & $\checkmark$ & $\checkmark$\\
        \cmidrule{2-4}
        \wue~(O) & 2019-04-17 16:22\textsuperscript{\ref{fig:colony_behavior_swarm_detected}} & $\checkmark$ & $\checkmark$\\
        \midrule
        \multirow{9}{*}{\jel~(S)} & 2015-05-06 18:02$^\star$ & $\checkmark$ & $\checkmark$\\
         & 2016-06-02 13:48$^\star$ & $\checkmark$ & $\checkmark$\\
         & 2016-05-30 10:03$^\star$ & $\checkmark$ & $\checkmark$\\
         & 2016-06-16 15:50$^\star$ & $\checkmark$ & $\checkmark$\\
         & 2016-06-01 13:20$^\star$ & $\checkmark$ & $\checkmark$\\
         & 2016-06-03 09:11$^\star$ & $\checkmark$ & $\checkmark$\\
         & 2016-06-13 03:30 & $\checkmark$ & $\checkmark$\\
         & 2016-06-16 10:52$^\star$ & $\checkmark$ & $\checkmark$\\
         & 2016-06-13 13:32$^\star$ & $\checkmark$ & $\checkmark$\\
         \midrule
        \multirow{2}{*}{\ind~(O)} & 2019-07-26 08:10 & $\checkmark$ & $\checkmark$\\
         & 2019-08-31 17:08\textsuperscript{\ref{fig:colony_behavior_varroa}} & $\checkmark$ & \\
        \bottomrule
    \end{tabular}
    }
    \label{tab:results_predictions}
\end{table}
This table only lists true positives of swarming events and false positives for comparison.
Swarms detected by apiarists on site are marked with~$^\star$.
\cref{fig:colony_behavior_swarm_real_detected} displays sensor traces of a typical swarming event.
%A prototypical swarming event is displayed in \cref{fig:colony_behavior_swarm_real_detected}.
All other swarming events were found by a combination of \algo\ and our approach: we ran \algo\ and examined the respective sensor readings to verify a swarming event.
Then we applied our AE model to the data and verified that it also found all events detected by \algo.
Additionally, we used our approach to find other anomalies or missed swarms.

The table states, that we found all true positives of swarming events, which can be seen in the groups of \textit{location (S)}.
Our AE found one additional swarm in \wue\ with T$_6$ and T$_8$, which is only found by \algo\ with T$_8$.
Furthermore, the groups of \textit{location (O)} list all anomalies detected as swarms by \algo\ with T$_6$, but are false positives of swarming events.

%Other anomalies are manifold and inherently hard to categorize, such as swarm-like events in temperature readings, contradicted by the weight sensor (\cref{fig:colony_behavior_swarm_detected}), and are thus not included in the table.
Other anomalies are manifold and inherently hard to categorize, such as excited bees due to outside influences, and are thus not included in the table.
Three exemplary non-behavioral anomalies are depicted in \cref{fig:colony_behavior2}.

%% file: sections/analysis.tex
\begin{figure}
    %\vspace*{-10pt}
    \centering
    \includegraphics[height=.9\textheight]{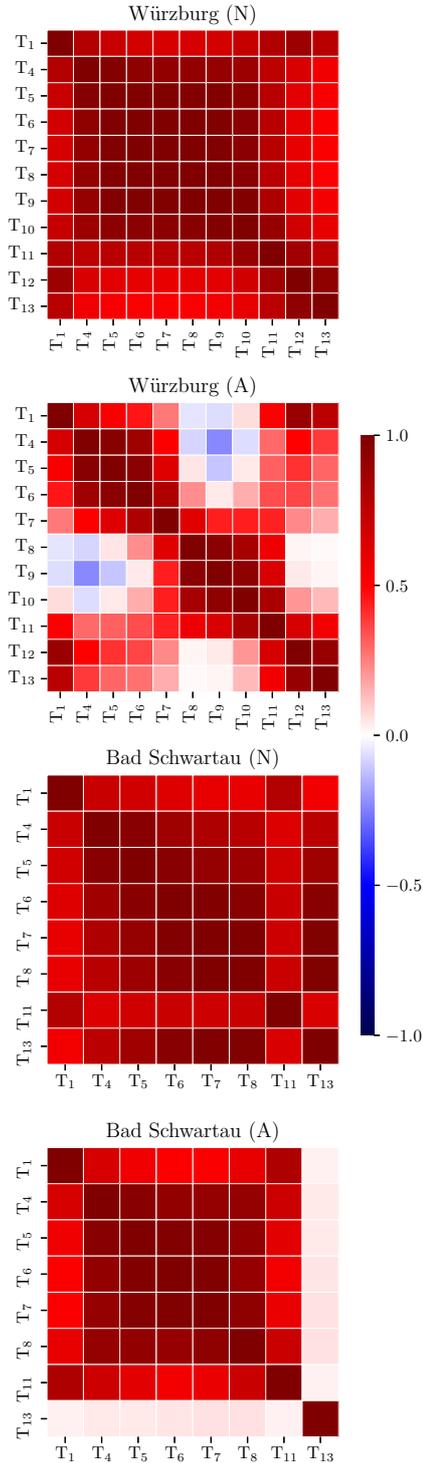}
    \caption{Sensor correlations. All figures display the Pearson correlation between temperature sensors within a given beehive. (N) stands for the dataset containing normal behavior and (A) for the dataset with anomalous behavior.}
    \label{fig:correlation_anomalies}
\end{figure}

\para{Analysis}
\cref{fig:correlation_anomalies} displays the inter-sensor correlation using the Pearson correlation coefficient.
When observing normal behavior, adjacent sensors correlate highly and positively.
Especially the sensors T$_4$--T$_{10}$, located centrally inside the hive, show high correlation between each other.
Sensors closer to the edges tend to correlate more with outside temperature sensors (T$_{12}$ and T$_{13}$).
%Especially the ``inner'' sensors (5-10, $r \geq 0.96$ and 5-8, $r \geq 0.88$ for \wue\ and \shw\, respectively), seem to build a cluster of high correlation.
%``Outer'' sensors only correlate with directly adjoining sensors as well as with their opposite counterpart.
Correlations during the anomalies are weaker, except between neighbors.
This confirms the findings in~\cite{zhu2019increase}, that certain sensor placements tend to capture swarms superiorly.

\begin{figure}
    \captionsetup[subfigure]{skip=-5pt}
    \centering
    \begin{subfigure}{\columnwidth}
        \centering
        \includegraphics[width=\textwidth]{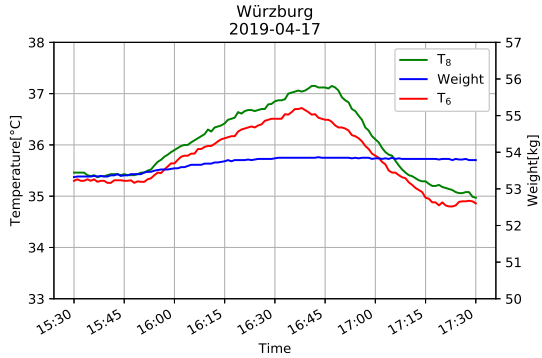}
        \caption{Swarm-like anomaly in sensors T$_6$ and T$_8$, but not within the measured weight.}
        \label{fig:colony_behavior_swarm_detected}
    \end{subfigure}%
    \vspace*{2pt}
    \begin{subfigure}{\columnwidth}
        \centering
        \includegraphics[width=\textwidth]{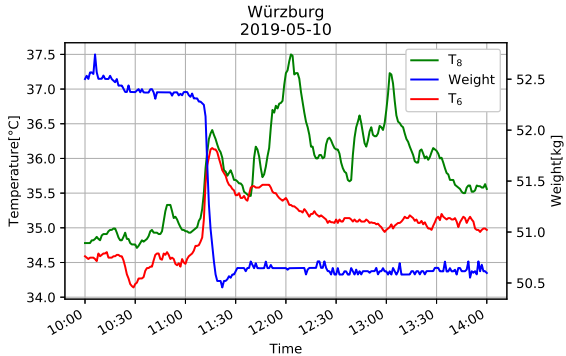}
        \caption{Swarm anomaly indicated by both T$_6$ and T$_8$, but additional swarms in T$_8$. Swarm anomaly within the weight.}
        \label{fig:colony_behavior_triple_swarm}
    \end{subfigure}%
    \vspace*{2pt}
    \begin{subfigure}[t]{\columnwidth}
        \centering
        \includegraphics[width=\textwidth]{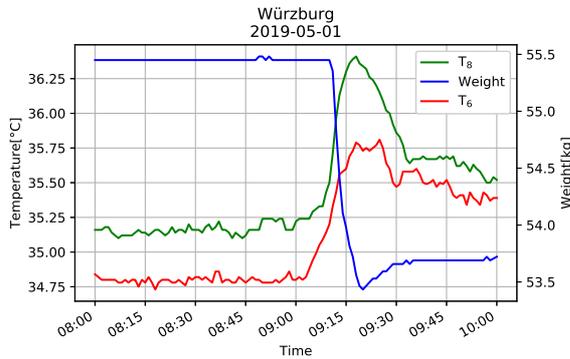}
        \caption{Swarm detected with T$_8$, but not with T$_6$ (\algo). Anomaly in both for AE. Swarm anomaly within the weight.}
        \label{fig:colony_behavior_swarm_not_detected}
    \end{subfigure}%
    \vspace*{2pt}
    \caption{Examples for behavior anomalies of swarm-like events.
%    All plots display sensor data of the weight sensor and two temperature sensors T$_6$ and T$_8$.
    % (\ref{fig:colony_behavior_swarm_not_detected}) Swarm anomaly only detected in one temperature sensor (T$_8$).
    % (\ref{fig:colony_behavior_triple_swarm}) Swarm anomaly with two subsequent, alleged after-swarms (T$_8$).
    % (\ref{fig:colony_behavior_swarm_detected}) Swarm-like anomaly (temperature) during normal behavior (weight).
    }
    \label{fig:colony_behavior}
\end{figure}

\begin{figure}
    \captionsetup[subfigure]{skip=-5pt}
    \centering
    \begin{subfigure}[t]{\columnwidth}
        \centering
        \includegraphics[width=\textwidth]{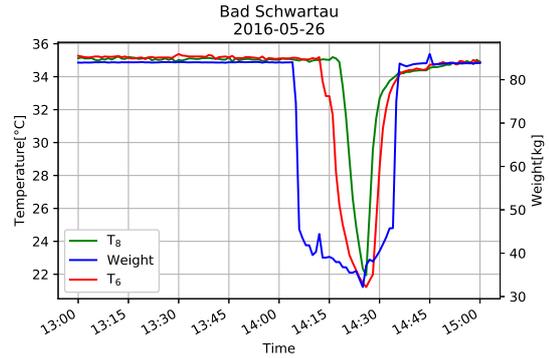}
        \caption{External interference of an opened apiary. The influx of outer air leads to the temperature drop.}
        \label{fig:colony_behavior_anomaly_open_hive}
    \end{subfigure}%
    \vspace*{2pt}
    \begin{subfigure}[t]{\columnwidth}
        \centering
        \includegraphics[width=\textwidth]{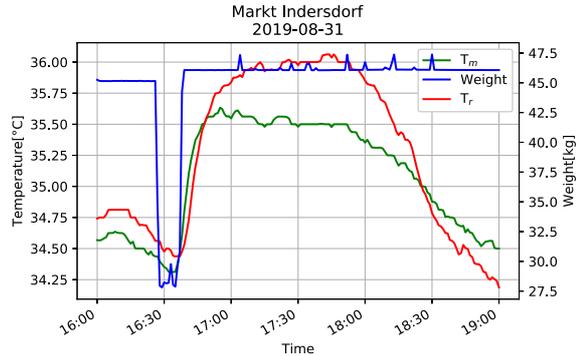}
        \caption{External interference by a possible varroa treatment. The beehive was opened, weight added, leading to the excitement of bees with a temperature increase. In contrast to our AE with T$_m$, \algo\ detected a swarm with T$_r$ and T$_m$.}% A swarm was triggered with T$_r$ and T$_m$ (\algo), but no trigger on T$_m$ with our approach.}
        \label{fig:colony_behavior_varroa}
    \end{subfigure}%
    \vspace*{2pt}
    \begin{subfigure}[t]{\columnwidth}
        \centering
        \includegraphics[width=\textwidth]{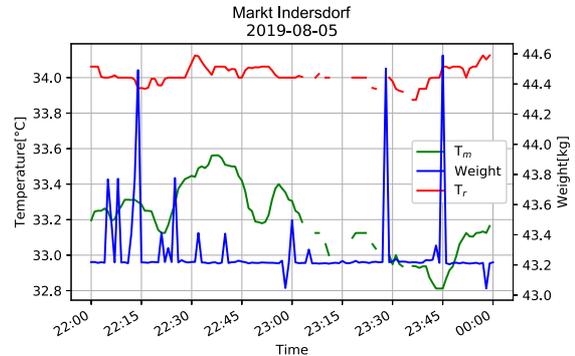}
        \caption{Sensor anomaly with missing values in T$_r$ and T$_m$, but not in the measured weights.}
        \label{fig:colony_behavior_sensor_defect}
    \end{subfigure}%
    \vspace*{2pt}
    \caption{Examples for external interferences (\ref{fig:colony_behavior_anomaly_open_hive}, \ref{fig:colony_behavior_varroa}) and sensor anomalies (\ref{fig:colony_behavior_sensor_defect}) that are present in the datasets.
%    All plots display sensor data of the weight sensor and two temperature sensors (T$_6$ and T$_8$).
    }
    \label{fig:colony_behavior2}
\end{figure}

\cref{fig:colony_behavior} shows swarm-like anomalies, \cref{fig:colony_behavior2} depicts sensor anomalies and external interferences.
%\cref{fig:colony_behavior,fig:colony_behavior2} show different types of colony behaviors.
All plots contain two temperature sensors (\hobos: T$_6$, T$_8$; \webee: T$_r$, T$_m$), as well as the measured weight.
%\cref{fig:colony_behavior_norm} displays normal behavior, as seen by the network during training.
%All other subfigures show different anomaly types or erroneously detected swarms.
As already mentioned in \cref{sec:experiments}, \cref{fig:colony_behavior_swarm_real_detected} shows a prototypical swarm as indicated by all three sensors and detected by both methods.

\cref{fig:colony_behavior_swarm_detected} shows an anomaly, that is falsely detected as a swarm when only looking at the temperature sensors.
The weight readings show normal behavior, thus contradicting the event trigger initiated by the temperature values.

On the other hand, \cref{fig:colony_behavior_triple_swarm} shows a colony swarm, as indicated by all three sensors.
%If we were to use only one temperature sensor (T$_8$), \algo\ would detect three swarms in this window, as would our method.
If we were to use only one temperature sensor (T$_8$), both methods would detect three swarms in this window.

%Using temperature sensor T$_6$, \algo\ does not classify the swarm displayed in \cref{fig:colony_behavior_swarm_not_detected} as such.
In contrast to \algo, AE detects the swarm displayed in \cref{fig:colony_behavior_swarm_not_detected} with temperature sensor T$_6$.
Only with temperature sensor T$_8$ find this swarm.

\cref{fig:colony_behavior_anomaly_open_hive} shows a window in which the beehive was opened, explaining the steep and quick drop in weight.
The temperature sensors trail this pattern with different delays, until the hive temperature has cooled to ambient temperature.
They quickly return to their initial readings as soon as the hive is closed again.

\cref{fig:colony_behavior_varroa} shows a varroa treatment with formic acid (hence the gain in weight). \algo\ detects swarms in this window for both temperature sensors, whereas our method only detects an anomaly in T$_r$.
That is possibly due to the fact, that sensor readings in T$_m$ are within one standard deviation of the training data.

\cref{tab:results_predictions} shows only swarm-like anomalies, but our method finds a lot more anomalies.
The larger portion of found anomalies are temperature readings well below \SI{30}{\celsius}.
Other monitoring anomalies, e.g. \cref{fig:colony_behavior_anomaly_open_hive}, are detected, too.

%% file: sections/discussion.tex
\para{Methodology}
%As described in \cref{sec:dataset}, both the definition of normal behavior and the method of determination, are highly error prone.
As described in \cref{sec:dataset}, the definition of normal beehavior is vague and the visual division is error prone.
%Hihihihihi
%We could either use a more rule based approach, i.e. all windows with sensor values drifting for more than two standard deviations, but that removes most swarming events from the test set.
%A clearer split of training and testing data can only be ensured by very thorough labeling of the sensor values, which has to be done on different sensors independently.

We selected the error threshold $\alpha$ introduced in \cref{sec:methods} manually, such that no normal behavior is detected as an anomaly in the validation set.
This approach allows to control the sensitivity of the AE.
It is a trade-off between fine-tuning for swarming detection and suppressing previously unknown anomalies, as seen in \cref{fig:colony_behavior_varroa}.
Methods that determine \(\alpha\) automatically also require labeled data.

In \cref{tab:results_predictions}, we listed all known swarms and their respective time of occurrence.
Due to the windowing technique described in \cref{sec:experiments}, we detect swarms at any position in their respective window.
% That is, anomalies are detected, whether they occur at the end of the window (predictive estimation) or at the beginning (historical estimation).
That is, anomalies are detected, both, at the end of the window (predictive estimation) or at the beginning (historical estimation).
% This predictive detection quality is highly dependent on the threshold, but enables apiarists to timely react to an ongoing or preeminent swarm.
This prediction quality is highly dependent on the threshold, but enables apiarists to timely react to an ongoing or preeminent swarm.

%% file: sections/conclusion.tex
\noindent
In this work we analyzed the possibilities of AEs in a new environment: bee colonies and their habitat.
% Our model was able to find more swarming events than \algo, a rule based method specifically designed for swarming detection.
Our model found more swarming events than \algo, a rule based method specifically designed for swarming detection.
%In this work we introduced a novel anomaly detection method for beehives.
% We showed that AEs are able to detect not only swarming events but also other anomalies.
Additionally, AEs detected not only swarming events but also other anomalies.
There are however several aspects with potential for improvements:

\para{Multivariate anomaly detection}
\hobos\ and \webee\ datasets enable us to use multivariate time series in contrast to the presented univariate temperature time series.
This allows us to refine predictions even further, as not only an anomaly within sensors is detectable, but an anomaly between sensors, i.e. inter- and intra-sensor anomalies of any type.
This could also minimize the overall error if only a subset of sensor shows abnormalities (e.g. in \cref{fig:colony_behavior_triple_swarm}).
% A coupled optical barrier serves as a digital counter which documents bees entering and leaving the hive.

\para{Method tweaking}
Instead of simply using the reconstruction error, we can adapt the loss to make different types of anomalies distinguishable.
For example, we could integrate the knowledge of temperature or weight patterns during swarming.

Another possibility to enhance anomaly detection, especially swarming detection, is to include a second training process to introduce $\alpha$ as a trainable parameter.
This requires a labeled dataset and is therefore subject to future analysis.

In future work we will experiment with other types of networks, e.g. generative models such as generative adversarial networks or variational autoencoders.
This has two key advantages:
\begin{enumerate*}
    \item[A)] they allow anomalies to be contained in the training set, and
    \item[B)] classification is based on probability rather than reconstruction error~\cite{an2015variational}.
    %\item[C)] they enable us to generate swarm-like events (hence generative).
\end{enumerate*}
The parameter \(\alpha\) would then be more interpretable.

\para{Hibernation period}
We excluded the months October through March in any dataset (cf. \cref{sec:dataset}).
Detecting anomalies during this hibernation time is subject to future work, as the assumption of a nearly constant temperature within the colony (\SI{34.5}{\celsius}) is void.
Especially in \shw, sea wind is an environmental influence that incurs very high deviations from the mentioned normal behavior which also increases the chances of sensor anomalies.
Additionally, internal temperature sensors start to mimic the patterns of the outside sensors.

\para{Dataset generation}
Due to its data-driven fashion, our method can be improved continuously by integrating collected information in \webee.
This project comprises a broad spatial distribution of apiaries, enabling us to collect a large amount of data fast.
Participating apiarists can further improve our model by labeling events presented to them.
Furthermore we can use our model as an alert-system to predictively warn beekeepers about ongoing anomalies, whose feedback can again improve our predictions.

%% file: sections/acknowledgements.tex
\noindent
% The scientific research and publications are supported by the we4bee project sponsored by the Audi Environmental Foundation.
This research was conducted in the we4bee project sponsored by the Audi Environmental Foundation.